\pdfoutput=1
\documentclass[letterpaper, 10pt, conference]{IEEEtran}
\IEEEoverridecommandlockouts

\usepackage{cite}
\usepackage{amsmath,amssymb,amsfonts}
\usepackage{graphicx}
\usepackage{booktabs}
\usepackage{multirow}
\usepackage{xcolor}
\usepackage{siunitx}
\usepackage{url}
\usepackage[hidelinks]{hyperref}
\usepackage[capitalize,nameinlink]{cleveref}

\makeatletter\newcommand{\figcaption}{\def\@captype{figure}\caption}\makeatother



\definecolor{linkblue}{HTML}{2F6FDB}  %
\begin{document}
\bstctlcite{IEEEexample:BSTcontrol}

\title{HITCH: Teaching a Vision-Language Model\\ to Operate a Humanoid through\\ a Whole-Body Harness}

\author{Qingzhou Lu$^{1,*}$, Chenxin Liu$^{1,*}$, Xinming Wang$^{2,*}$, Liu Cao$^{1}$, Yanjiang Guo$^{1}$, Mengdi Xu$^{1}$,
and Jianyu Chen$^{1,3}$\\[2pt]
{\normalsize Project website: \textcolor{linkblue}{\url{https://hitch.axell.top}}}}

\twocolumn[{%
  \csname @twocolumnfalse\endcsname
  \maketitle
  \begin{center}
    \includegraphics[width=\textwidth]{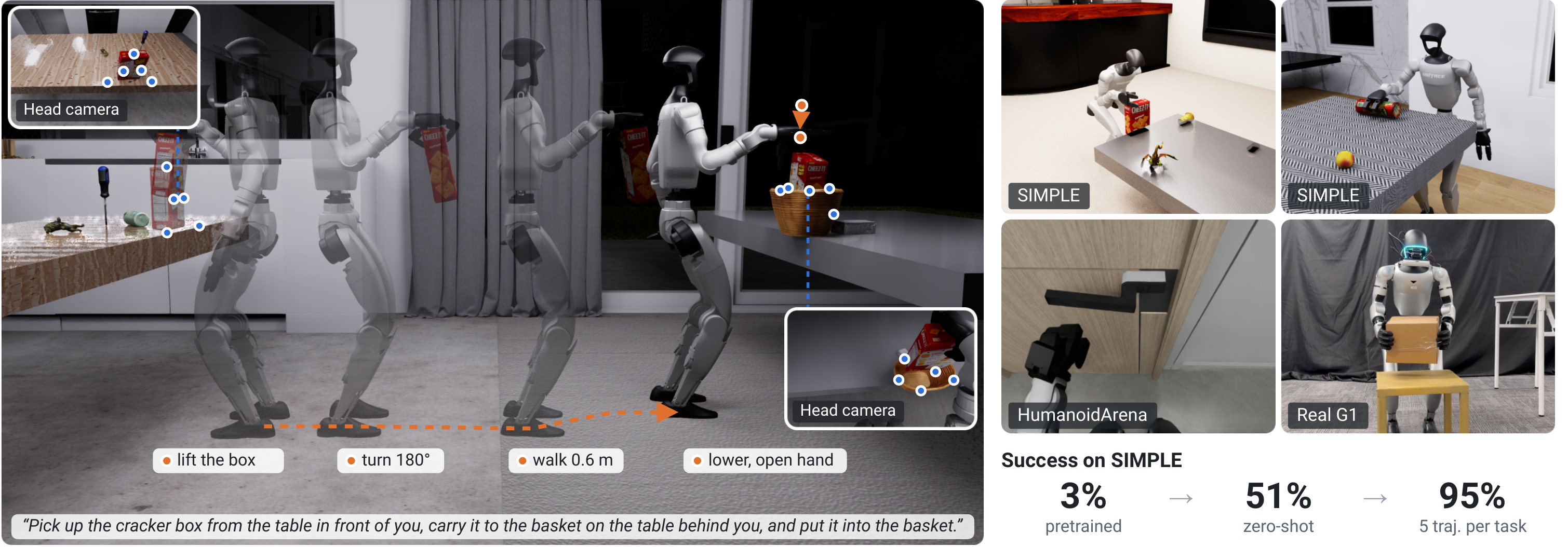}
    \figcaption{A VLM fine-tuned on teacher trajectories controls a G1 humanoid through a harness. Left: one
      evaluation episode of SIMPLE, with the robot drawn at several moments. The agent measures the box and the
      basket by choosing points in its head-camera images (insets) and receives their 3D positions (blue). It then
      commands the body: it lifts the box, turns, walks to the other table, and lowers its hand into the basket
      through waypoints (orange). Right: other tasks in SIMPLE, HumanoidArena, and on a real G1,
      and the success rate on SIMPLE of the pretrained model, of the model trained on the base tasks only, and of
      the model trained on the base tasks and 5 trajectories per SIMPLE task.}
    \label{fig:teaser}
  \end{center}
  \vspace{1em}
}]
{\renewcommand{\thefootnote}{}\footnotetext{$^{*}$Equal contribution. $^{1}$Tsinghua University.
$^{2}$Institute of Automation, Chinese Academy of Sciences. $^{3}$Shanghai Qizhi Institute.}}
\thispagestyle{empty}
\pagestyle{empty}

\begin{abstract}
Enabling vision-language models (VLMs) to control humanoid robots requires both reliable whole-body interfaces and
knowledge of how to use the body that carries over across tasks. We present HITCH (Harness Interfaces To Command
Humanoids), a harness in which perception returns the 3D position of any point the VLM selects in its image, and
execution turns commands for the pelvis, torso, hands, walking, and gaze into whole-body motion with a learned tracker.
We find that both measuring the scene and executing commands accurately raise task success. We then fine-tune a VLM on trajectories in which
teacher policies solve five base tasks through HITCH, so that it learns how to operate the humanoid. Without
task-specific trajectories, the fine-tuned model completes 51\% of SIMPLE episodes and 42.5\% of HumanoidArena episodes,
and adding five trajectories per SIMPLE task raises its SIMPLE success to 95\%. On a real G1, it carries a box and walks
to a goal without real training data.

\end{abstract}

\section{Introduction}
\label{sec:intro}

General-purpose humanoid robots should carry out everyday loco-manipulation tasks from language instructions, for
example, walking to a table, crouching to pick up a cup, and placing it in a basket on another table. Each of these
steps requires the legs, torso, and arms to act together while the head camera keeps the target in view, and every
motion must match the geometry of a scene the robot has never seen.

A common approach trains vision-language-action (VLA)
models~\cite{wei2026psi0,shao2026holom} or world action
models~\cite{yue2026beingm07} that map observations and instructions directly to whole-body actions.
These models produce fluent motion, but they usually need many demonstrations of each task, and their performance
can drop when only a few are available~\cite{wei2026simple}. For a humanoid, each demonstration requires
whole-body teleoperation or motion planning, which makes every new task expensive to add.

Agents built on foundation models offer another route, but on a humanoid they lack two things. Prompted language
models and vision-language models (VLMs) already solve many tabletop tasks without task-specific
training~\cite{liang2022codepolicies,huang2023voxposer,huang2024rekep}. The first thing they lack on a humanoid is an
interface to the whole body. Existing humanoid agents mostly select skills from a library or command
task-space controllers~\cite{yuan2025being0,ren2026cybowaiter}, which leave the posture of the legs and torso to the
controller. A task may require the agent to set this posture itself, and the interface must execute each command accurately, since the agent observes again only after a
command finishes. The second is skill in operating a humanoid body. Pretraining contains few examples of humanoid
motion: a model may understand 3D space well yet know little about how far a humanoid can reach, or when it must step
or crouch to bring its hands to a target. RAX acquires such knowledge through online exploration of each new task,
with success that grows with the exploration budget~\cite{huang2026rax}. Yet much of this knowledge is shared
across tasks and could be learned once.

We first build HITCH (Harness Interfaces To Command Humanoids), a harness through which a VLM observes and
controls the robot. Its perception interface lets the
model measure the scene: it selects a point in the image and receives the 3D position of that point from depth, so it
does not have to guess distances from the image alone. Its execution interface exposes commands for each body part,
such as pelvis height, torso orientation, hand waypoints, walking, and gaze. A planner converts them into whole-body
references, which a learned tracker executes on a Unitree G1. We train this tracker further on the planner's
references. Given the same commands, it executes them more accurately than baseline trackers, and more tasks succeed
with it.

With this harness, the VLM only has to choose the next call, but nothing in the harness tells it how to choose. We therefore fine-tune Qwen3.8-27B on trajectories in which teacher
policies use the same interfaces to solve five base tasks: tabletop pick-and-place, carrying objects between pieces
of furniture, navigation, opening drawers and cabinet doors, and opening room doors. From these trajectories, the
model learns skills for operating a humanoid through the harness, such as where to stand, how low to crouch, how far
to lean, and what to measure before acting.

We find that these skills carry over to new tasks in simulated
benchmarks~\cite{wei2026simple,wang2026humanoidarena}, and to real scenes on a G1 without real training data. On the SIMPLE humanoid benchmark, the model
trained on the base tasks completes about half of the episodes without task-specific trajectories, and adding
five trajectories per task raises its success to 95\%. Compared with training on the task trajectories
alone, training on the base tasks as well raises success at every number of trajectories we tried. Without its perception interface, the model trained on the base tasks solves only
2.8\% of SIMPLE's episodes. On a real G1, the model completes 13 of 20 carrying and 6 of 10 navigation episodes.

Our contributions are:
\begin{itemize}
  \item a harness for humanoid agents with two interfaces: perception, which measures the scene on demand, and
        execution, which turns body-part commands into accurate whole-body motion;
  \item teacher policies that solve five humanoid base tasks through these interfaces, acting only on what the agent
        observes, and a VLM fine-tuned on their trajectories that carries over to new tasks;
  \item experiments on two humanoid benchmarks and a real G1, covering transfer to new tasks, learning from a few
        trajectories per task, transfer to real scenes, and ablations of perception and execution.
\end{itemize}

\section{Related Work}
\label{sec:related}

\begin{figure*}[t]
  \centering
  \includegraphics[width=\textwidth]{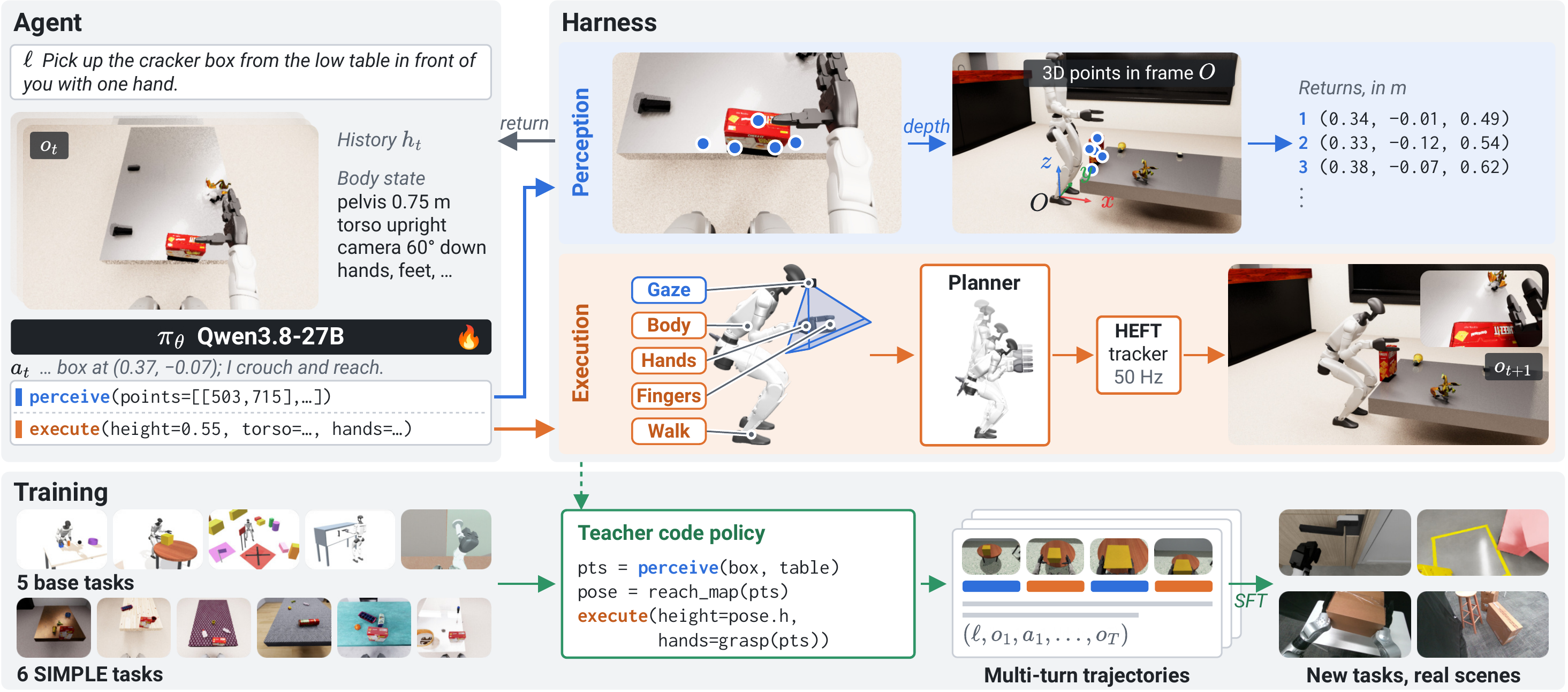}
  \caption{Overview. Top left: in each turn, the agent $\pi_\theta$ (Qwen3.8-27B) receives the instruction $\ell$ and
    the history $h_t$ of head-camera images and body states, and writes a response $a_t$ that contains one call. Top
    right: the call goes to the harness. A perception call selects points in the image and returns their 3D positions
    in the observation frame $O$, measured from depth, together with the marked image. An execution call sets the body,
    hands, fingers, walking, and gaze; a planner turns it into a whole-body reference, which the HEFT tracker follows
    at \SI{50}{Hz}, and the call returns the next observation. Bottom: teacher
    code policies act through the same harness on five base tasks and six SIMPLE tasks. The
    agent is fine-tuned on their multi-turn trajectories and evaluated on new tasks without or with a few task
    trajectories, and on a real G1 in new scenes.}
  \label{fig:method}
\end{figure*}

\subsection{Humanoid Whole-Body Tracker}
The low-level controller is a central component of a humanoid system, as it largely determines the precision and
robustness of the executed motion. With recent advances in motion tracking and teleoperation, such controllers are now
mostly learned. Whole-body trackers trained with reinforcement learning on human motion follow full-body kinematic
references~\cite{he2024omnih2o,ji2024exbody2,chen2025gmt,liao2025beyondmimic,ze2025twist2,chen2026holomotion,liu2026heft},
and SONIC~\cite{luo2025sonic} scales this approach to one controller that can be driven by teleoperation, a
kinematic planner, or a VLA. Decoupled designs pair a learned lower-body policy with directly commanded
arms~\cite{li2025amo,ben2025homie}, and HOVER~\cite{he2024hover} supports several command modes in one
policy. Since a tracker reduces whole-body control to producing a feasible reference, it can execute the output of a
higher-level module, such as these planners or a grasp and motion planning pipeline~\cite{dong2026hero}. We extend this design to serve an agent: a planner converts the agent's body-part commands into whole-body
references, and a tracker trained with HEFT~\cite{liu2026heft} executes them. We further train the tracker on
references produced by this planner so that it follows the agent's commands more closely, and compare it with
other trackers as executors of the same commands.

\subsection{Vision-Language-Action and World Action Models}
Vision-language-action (VLA) models build on pretrained VLMs to map images and instructions to
actions~\cite{kim2024openvla,black2024pi0,physicalintelligence2025pi05}, and world action models predict future
observations together with actions~\cite{ye2026dreamzero,nvidia2026cosmos3}. Humanoid models of this kind are either
pretrained on human or cross-embodiment data and fine-tuned on whole-body
demonstrations~\cite{wei2026psi0,shao2026holom,xu2026lambda0,yue2026beingm07}, or
trained to predict locomotion, posture, or motion commands for a lower-level
controller~\cite{xue2025leverb,cheng2024navila,bai2026egoactor}. Benchmarks such as HumanoidArena
and SIMPLE evaluate them on shared tasks~\cite{wang2026humanoidarena,wei2026simple}.

Because the humanoid models above learn
motion directly from demonstrations, their generalization is often limited to variations of the trained tasks. We take a different approach: our agent leaves
motion to the tracker and learns how to act through interfaces, and what it learns transfers to unseen tasks with
little or no task-specific data.

\subsection{Foundation Models and Agentic Robot Control}
Language and vision-language models can control robots without task-specific training. They select
skills~\cite{ahn2022saycan,huang2022innermonologue} or write policy
code~\cite{liang2022codepolicies,huang2023voxposer}, and solve many manipulation tasks this way. Most of this progress
concerns arms and mobile manipulators; on humanoids, foundation models are mostly prompted to drive skill libraries or task-space
controllers~\cite{yuan2025being0,ren2026cybowaiter,yang2026handoff}. When a model lacks the knowledge a
task requires, it can gain it from experience: RAX explores at scale and compiles successful attempts into reusable
programs~\cite{huang2026rax}, but each new task still requires online exploration. Fine-tuning offers another route, and open VLMs
trained on trajectories collected in a harness, which defines the observations, tools, and feedback of the agent,
learn to operate robot arms~\cite{liu2026guava,chen2026showharness}.

A humanoid poses a harder problem for agents: its hands reach a target only after it walks, crouches, or turns its
torso, and its camera moves with its head, so the agent must learn how to operate the whole body. Our agent
measures the scene and commands each body part through separate interfaces, and learns these skills from teacher
trajectories on base tasks so that they carry over to new tasks.

\section{Method}
\label{sec:method}

Our agent is a VLM that follows a language instruction on a Unitree G1 humanoid over a sequence of turns
(\cref{fig:method}). In each turn, it makes one call: a query to a perception interface, which measures the scene, or
a command to an execution interface, which moves the body. Together, the two interfaces form the harness through
which the VLM observes and controls the robot. The agent learns to operate the robot through these interfaces from
the trajectories of teacher policies. \Cref{sec:setting} defines the agent loop and two properties of this setting that
shape our design, \cref{sec:execute,sec:perceive} describe the two interfaces, \cref{sec:teacher} how we collect teacher
trajectories, and \cref{sec:training} how we fine-tune the agent on their trajectories.

\subsection{Problem Setting}
\label{sec:setting}
We model an episode as a multi-turn interaction between the VLM and the robot. It starts from a language instruction
$\ell$ and an initial observation $o_1$. At turn $t$, the agent produces a response $a_t$
that contains one interface call and may contain free-form reasoning, and the call returns the next observation:
\begin{equation}
  a_t \sim \pi_\theta(\cdot \mid \ell, h_t),
  \qquad h_t = (o_1, a_1, o_2, \ldots, a_{t-1}, o_t),
\end{equation}
where $\pi_\theta$ is the VLM and $h_t$ is the history. The agent conditions on the whole interaction so far, so it
can refer back to positions it measured in earlier turns.

A perception query returns the measured positions and the image with the queried points marked. A command returns a
new head-camera image and the body state after the motion, covering the pelvis, torso, feet, palms, fingers, and
camera. Each item of the body state gives both the value the agent commanded and the value read from joint encoders
and the IMU, so the agent can detect when a hand was blocked by an object or a crouch fell short. In a typical
episode, the agent looks for the target, measures it, moves, and looks again to check the result, repeating this
cycle until the task is complete.

Two properties of this setting shape our design. First, the agent decides at a low rate. A VLA observes again
after each short chunk of actions and adjusts its motion as it goes, whereas our agent spends seconds on each turn,
and a command runs to completion before the agent observes again. The agent must therefore know where its targets
are before it acts: if it could only judge positions from the image, it would have to home in on a target over
several turns, moving, looking, and adjusting by the offset it sees. Its commands must also be executed
accurately, with the body reaching the commanded posture and the palms their commanded positions. The
next observation reveals an error, but correcting it costs another turn, and some errors cannot be undone, such as
knocking the object off the table.

Second, every decision in the training data must be explainable from what the agent receives: the instruction and
its observations. Our teachers are code policies in simulation and could read the pose of any object from the
simulator, which the agent cannot. Suppose the robot faces a table and must put an object into a basket on a second
table behind it. If the instruction does not mention the basket and the teacher turns straight to it, nothing in
the agent's input explains the turn. A policy trained on such demonstrations can reproduce it only by remembering
the training layouts, and in a new scene it may place the object at the remembered position without having looked.
This is a known failure of imitation when the teacher acts on information the student
lacks~\cite{weihs2020advisor}. The same holds for facts that are constant across a task, such as the side on which
the destination lies or how the object must be left. A policy trained on many demonstrations of a task picks up
such facts from the demonstrations, so a short instruction suffices for it. An agent that has not seen these
demonstrations must be told such facts in the instruction.

The first property shapes the two interfaces: the perception interface (\cref{sec:perceive}) lets the agent
measure the positions of its targets before it acts, and the execution interface (\cref{sec:execute}) turns each
command into accurate whole-body motion. The second shapes the teachers (\cref{sec:teacher}), which act only on the
instruction and their own measurements, and the instructions we write for benchmark tasks (\cref{sec:exp-setup}),
which state the facts that hold in every episode of a task but no positions or other facts that vary between
episodes.

\subsection{Whole-Body Execution}
\label{sec:execute}
\begin{figure}[t]
  \centering
  \includegraphics[width=\linewidth]{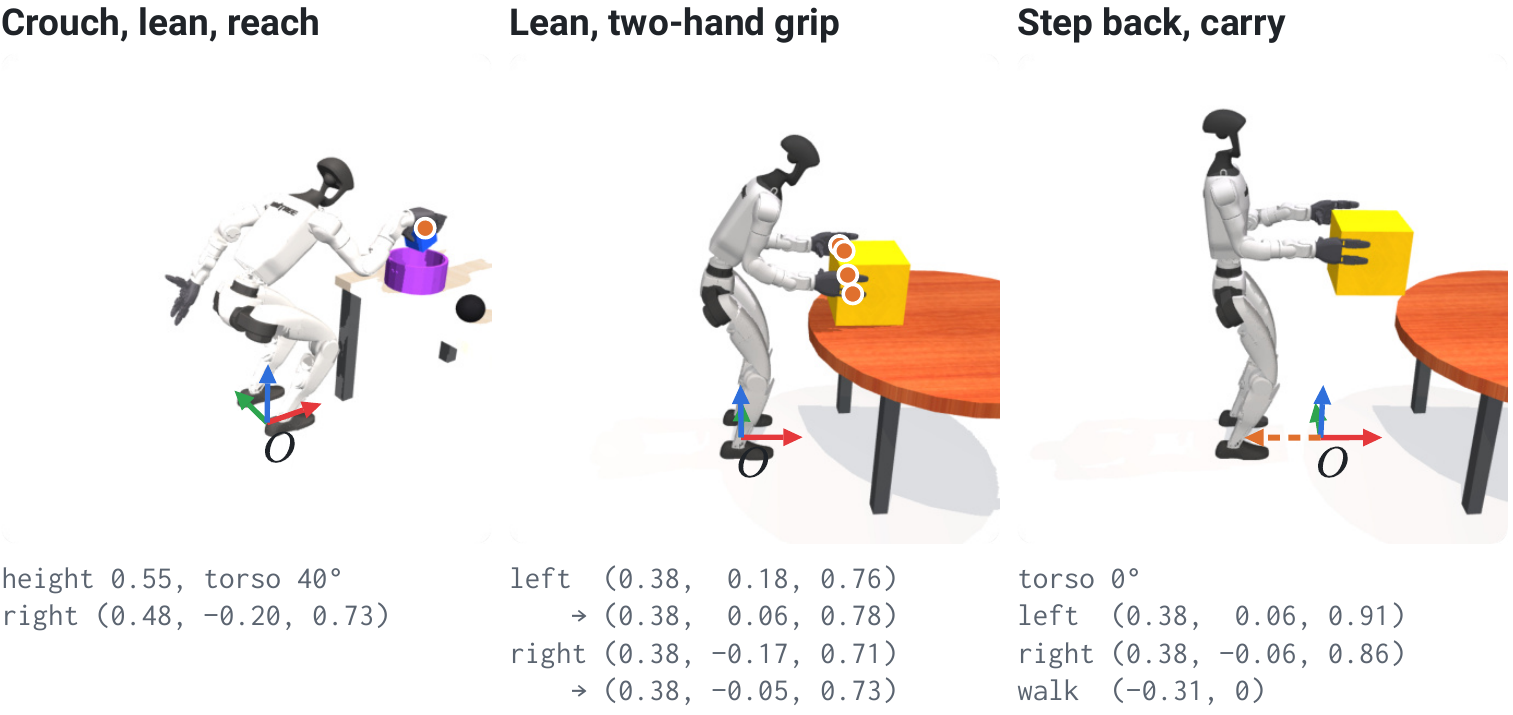}
  \caption{Postures set by single commands in teacher episodes of the base tasks. A command gives positions in the
    observation frame $O$ of the latest observation (axes), in metres. Orange dots mark the palm positions that the
    left and middle commands set; the dashed arrow on the right marks the walk target. Left: crouch, lean forward,
    and place a cube into a bowl. Middle: grip a box with both hands, each palm moving through two waypoints, while
    keeping the \SI{40}{\degree} lean of the previous command. Right: lift the box and step back. Below each frame
    are the fields of its command.}
  \label{fig:execution}
\end{figure}

The execution interface must expose the degrees of freedom that our tasks use, including pelvis height and torso
pitch and yaw, and execute each command as stated. A planner converts each command into a whole-body
reference, and a learned tracker follows this reference.

\subsubsection{Observation Frame}
\label{sec:frame}
Commands and measurements use the observation frame $O$. Its origin lies on the ground below the pelvis, with $x$
along the pelvis heading, $y$ to the left, and $z$ up. $O$ is redefined at every observation and stays fixed while a
command runs, so positions measured since the latest observation apply directly to the next command. If the robot
has not walked since an earlier observation, the new observation gives the pose of the new $O$ in the old one, so
earlier positions can be converted. After a walk, the
robot's displacement is known only approximately, so this pose is uncertain and earlier positions are no longer
reliable.

\subsubsection{Tracker}
\label{sec:tracker}
The tracker is a whole-body motion tracking policy trained with HEFT~\cite{liu2026heft}. It is trained for a G1 with
either Dex3 hands or rubber hands, runs at \SI{50}{Hz}, follows a reference for all joints, and keeps the robot
balanced. We further train it on references that the planner produced while teachers solved the base tasks
(\cref{sec:teacher}). Some of the postures the agent commands are rarely covered by human motion data, and training
on them improves tracking accuracy and avoids unneeded compensating motion. We also apply random forces to the torso
and hands, which makes the tracker more robust and lets the hands exert the forces that tasks require, for example
when opening a door or holding an object. Given the same commands, the resulting tracker executes them more
accurately than baseline trackers, and more tasks succeed with it (\cref{sec:exp-execution}).

\subsubsection{Planner}
\label{sec:planner}
A command states what each body part should do through the following fields, which are executed in the order
listed. \textbf{Body:} the pelvis height between \SI{0.30}{m} and \SI{0.79}{m}, and the torso pitch and yaw
relative to a default posture for that height. \textbf{Hands:} for each hand, a sequence of waypoints for the palm
frame in $O$, each with a position, an orientation, and a duration; while the body changes height or leans, a hand
can instead stay fixed in space or move with the pelvis or the torso. \textbf{Fingers:} close or open, on dexterous
hands. \textbf{Walking:} to a target pose in $O$ or along given footsteps. \textbf{Gaze:} the yaw and pitch of the
head camera, set last so that the next observation is taken from the requested view. Because one command sets all
body parts, the agent can compose postures such as a crouch with a forward lean and a two-hand reach (\cref{fig:execution}). The system
prompt names the robot model and the fields it supports, defines the palm frame, and gives the hand geometry that
the agent needs to place the palm.

The planner solves inverse kinematics for the commanded pelvis, torso, and hand targets under joint, velocity, and
self-collision limits. It adds footstep motion for walking and streams the result to the tracker as one continuous
reference.

The planner covers walking on flat ground, crouching, leaning, reaching, and grasping. Other motions can be added
by producing their references for the same tracker, for example with a motion generation model.

\subsection{Active Perception}
\label{sec:perceive}
The perception interface lets the agent measure the positions of the objects and surfaces it will act on before it
acts. VLMs without dedicated spatial training are poor at estimating metric positions from an
image~\cite{chen2024spatialvlm,cheng2024spatialrgpt}, whereas VLMs trained on pointing data localize image points
accurately~\cite{deitke2024molmo,zhou2025roborefer}. The interface splits the work accordingly: the agent chooses
where to look and which points to measure, and depth supplies their positions.

\begin{figure}[t]
  \centering
  \includegraphics[width=\linewidth]{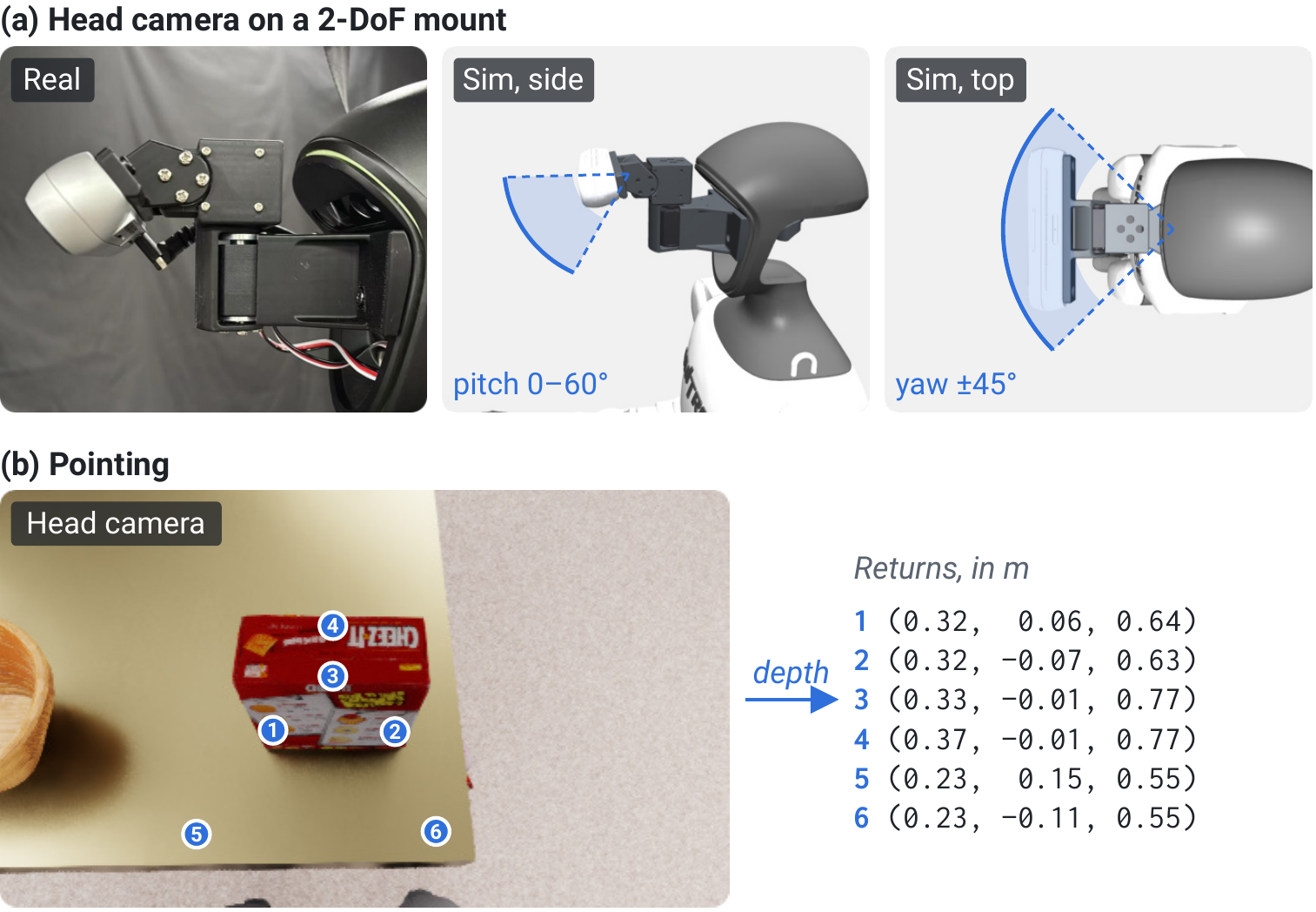}
  \caption{Active perception. (a) The D455 on a 2-DoF mount on the G1's head, on the real robot and in simulation,
    seen from the side and from above. The mount turns the view from level to $60^\circ$ down and up to $45^\circ$
    either way. (b) A perception query of the agent in SIMPLE: it selects six points on the box and the table in the
    head-camera image and receives their positions in $O$.}
  \label{fig:perception}
\end{figure}

To let the agent look for what it needs, we add a 2-DoF mount on top of the G1's fixed head and
place an Intel RealSense D455 on it (\cref{fig:perception}a), and reproduce its geometry in simulation. Every command sets
the gaze of the camera, with yaw within $\pm45^\circ$ and pitch from level to $60^\circ$ downward. Since the agent observes nothing outside the camera's view, it has to look down at the
table in front of it, turn toward a destination at its side, or look ahead before walking.

On the real robot and in simulation, the agent receives a $640\times400$ RGB image with a field of view of
$90^\circ\times64^\circ$ (horizontal by vertical), together with an aligned depth map. On the real robot, the D455 captures RGB at $1280\times800$ and depth
at $640\times360$; the RealSense SDK aligns the depth to the RGB image, and we downsample both to $640\times400$. Raw
D455 depth is noisy and has holes, so we refine each frame with LingBot-Depth~\cite{tan2026lingbotdepth}, a depth
completion model. We calibrate the camera on its mount with a ChArUco board held in the robot's hand, and for
each commanded gaze we solve for the servo angles that align the camera's optical axis with the one in simulation.
We configure the cameras in our simulation and in both benchmarks to this setup. In the benchmarks,
this camera replaces the default ones, which are fixed to the head and have wider fields of view, about
$110^\circ\times78^\circ$ in SIMPLE and $106^\circ\times89^\circ$ in HumanoidArena.

A perception query turns points that the agent selects in the latest image into positions in $O$
(\cref{fig:perception}b). Each point is given as integer coordinates normalized to $[0, 1000]$, the convention that
Qwen3-VL uses for grounding~\cite{bai2025qwen3vl}. The interface reads the depth at that pixel and back-projects it
into $O$ with the camera pose at the time of capture, correcting for lens distortion on the real robot. It reports
points without depth as invalid and flags points whose $5\times5$ neighborhood has a large depth spread, which usually
indicates an object edge. The returned image marks each point, so the agent can check whether a point landed on the
intended surface and point again if it did not.

\subsection{Collecting Teacher Trajectories}
\label{sec:teacher}
The interfaces define what the agent can do, but a pretrained VLM knows little about how a humanoid should use them.
We teach the agent to use them through trajectories of teacher policies, code policies that solve tasks in
simulation through the same interfaces as the agent.

\subsubsection{Tasks}
Five base tasks provide the training data shared by all tasks (\cref{fig:method}, bottom): picking and placing
objects with one hand on low and high tables, carrying a box with both hands between pieces of furniture, walking to
a goal around obstacles, opening or closing drawers and cabinet doors, and pushing or pulling open a room door with a
lever handle. Within each base task, we randomize the objects and their placement, the destinations, the furniture
layout, the robot's starting pose, and the appearance of the scene. Steps such as approaching a table, crouching,
and measuring a grasp point recur across the base tasks, and we expect an agent that learns these steps to reuse them in new tasks that combine them differently.

Six tasks of the SIMPLE benchmark serve as new tasks that the agent learns from a few trajectories. For each
task, our teachers solve the scene layouts of 50 of the demonstrations that SIMPLE releases, and we evaluate
on SIMPLE's released evaluation episodes.

\subsubsection{Teacher Policies}
The teachers measure where things are before they act. When a target is not in view, for example, a teacher first looks
where the instruction points. The teacher chooses its posture and where to stand from a precomputed
map of what the hands reach in each posture. The map depends only on the robot's body, and the teacher applies it to
positions it has measured, so each choice is still a function of the agent's observations. The agent never sees the
map and has to learn these choices from the teachers' actions.

To enforce this behavior, we require that every action of a teacher can be derived from its own observations and
reasoning, which it writes in its response. The teacher receives only what the instruction states and computes every position it acts on from its measurements and the robot's body state.

We wrote the teachers with Claude Opus~5.5. Each teacher episode is recorded as a multi-turn trajectory with the
same system prompt and observations that the agent receives, with the wording of each response sampled from
equivalent phrasings, and we keep only successful episodes.

\subsection{Training}
\label{sec:training}
We fine-tune Qwen3.8-27B on the teacher trajectories. Each trajectory forms one multi-turn sample.
Each response holds one call and may hold the teacher's reasoning. We update all
weights of the language model, keep the vision encoder frozen, and apply the loss only to the teacher's responses,
\begin{equation}
  \mathcal{L}(\theta) = -\sum_{t} \log \pi_\theta(a_t \mid \ell, h_t),
\end{equation}
summed over the turns of each trajectory.

The base-task model is trained on the five base tasks, with 500 trajectories each. To learn a new
task, we train on its few trajectories either alone or mixed into the base task data; both start from the pretrained
VLM. The mixture lets the new task build on
the skills of the base tasks.

We train every model in two versions, with and without the teacher's reasoning in the responses, to study how
reasoning affects an agent that controls a humanoid through our harness. The version without reasoning also acts
without it at test time.

\section{Experiments}
\label{sec:experiments}

We first test whether the execution interface carries out commands accurately enough for the tasks. We then
evaluate the agent: whether training on the base tasks lets it solve benchmark tasks without task-specific
trajectories, how quickly it learns a new task from a few trajectories, how much it relies on perception, and how
reasoning in its responses affects it. Finally, we run the agent on a real G1.

\begin{figure*}[t]
  \centering
  \includegraphics[width=\textwidth]{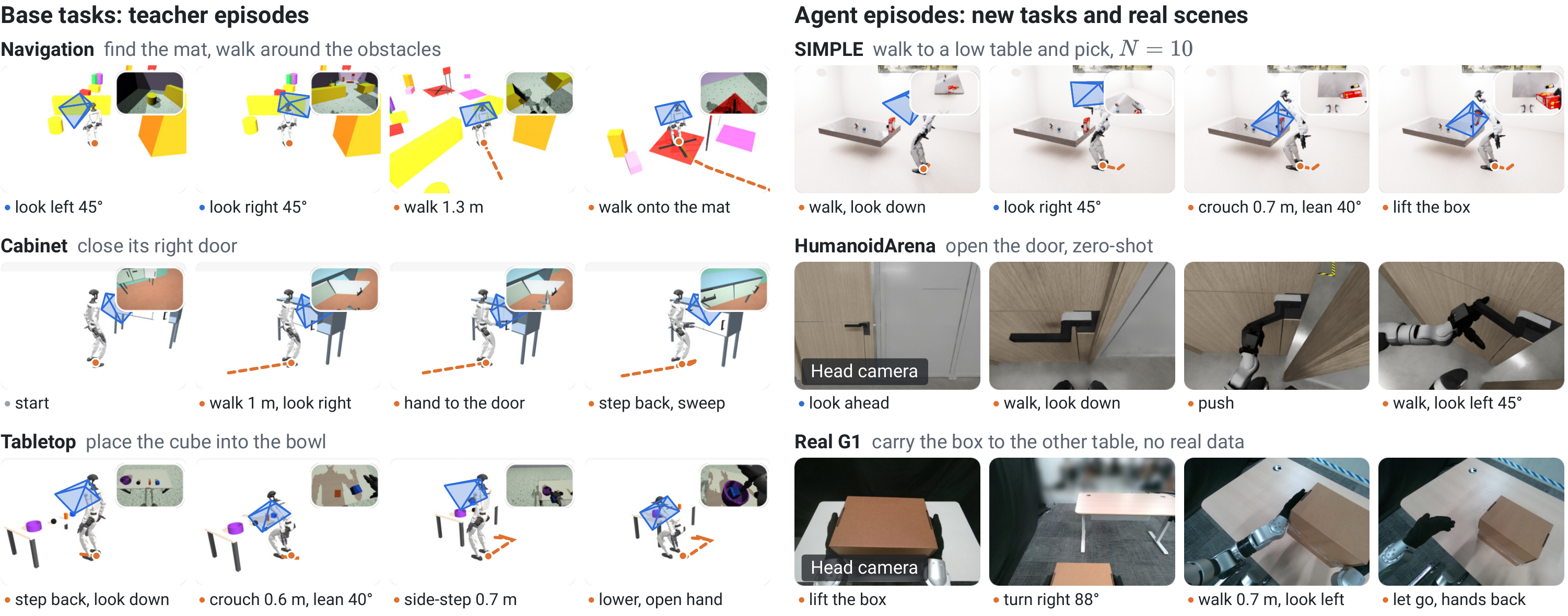}
  \caption{Episodes in time order, one row each, with the call that returned each frame below it. Left: teacher
    episodes of three base tasks. Right: the agent on SIMPLE after training on the base tasks and ten SIMPLE
    trajectories per task, and on HumanoidArena and the real G1 after training on the base tasks only. Third-person
    frames show the head-camera image (top right), the head camera's view (blue), and the path walked so far
    (orange); the HumanoidArena and real G1 rows show head-camera images.}
  \label{fig:filmstrips}
\end{figure*}

\subsection{Experimental Setup}
\label{sec:exp-setup}
We test the agent mainly on two humanoid benchmarks, SIMPLE~\cite{wei2026simple} and
HumanoidArena~\cite{wang2026humanoidarena}, which play different roles. On SIMPLE, we measure success both without
task-specific trajectories and after training on a few trajectories of its tasks. HumanoidArena is held out: no model is
trained on its tasks. We keep the scenes, objects, physics, success checks, and evaluation episodes of both
benchmarks, and change only what our harness requires.

\textbf{Benchmarks.} For SIMPLE, we wrote teachers for six core tasks and evaluate each on its 30
released evaluation episodes. From HumanoidArena, we use four tasks: navigation, boxing, opening a door, and moving a
box onto a shelf, each with 20 official episodes. The remaining tasks require motions that our planner does not yet
support, such as sitting, kicking a ball, or stepping over obstacles. \Cref{fig:filmstrips} shows episodes from both benchmarks.

\textbf{Adapting the benchmarks.} Our planner and tracker drive the G1 in both benchmarks, and we make two further
changes. First, our head camera on its 2-DoF mount replaces the fixed head camera of each benchmark, so the agent
observes the scene as in training and on the real robot. Second, we rewrite the official instructions in the wording
of our training data. Each instruction states the facts that hold in every episode of the task.
For example, SIMPLE's instruction for carrying an object between tables calls them ``table1'' and ``table2'', while
ours says that the basket is on the table behind the robot; HumanoidArena's ``Move the box from the table onto the
shelf'' does not say where the shelf is, while ours places it on the robot's left. No instruction gives positions,
and each is the same for every model and episode.

\textbf{Metrics.} We report the success rate. An episode succeeds if the benchmark's check passes within \SI{45}{s}
of simulated time in SIMPLE and \SI{60}{s} in HumanoidArena. These limits leave room for our agent, whose
motions through the harness are slower than those of a policy acting at high frequency, and which often moves to get
a better view and confirm its targets before it commits to a motion.
HumanoidArena's tasks involve more walking, so we give them more time. Each SIMPLE result covers 180 episodes and each HumanoidArena result 80. Successful episodes of the fine-tuned agents without reasoning
take 15 to 21\,s on average in SIMPLE and 21 to 27\,s in HumanoidArena.

\textbf{Models.} We compare the pretrained Qwen3.8-27B in our harness with models fine-tuned on three kinds of data:
the base tasks only, $N$ SIMPLE trajectories per task only, and the base tasks together with $N$ SIMPLE trajectories
per task, with $N$ = 5, 10, or 30. Each fine-tuned model is trained with and without reasoning, and the pretrained
model is run with and without its thinking mode. We also run the much larger Qwen3.8-Max through its API, without
fine-tuning, on SIMPLE. Unless noted, we report the models without reasoning, and
\cref{sec:exp-reasoning} compares the two.

\subsection{Execution}
\label{sec:exp-execution}
\begin{table}[t]
  \caption{Execution. Hand error (cm), walk error (\%), and teacher success rate (\%).}
  \label{tab:execution}
  \centering
  \setlength{\tabcolsep}{4.5pt}
  \begin{tabular}{lccccc}
    \toprule
    & \multicolumn{4}{c}{Tracking error} & \\
    \cmidrule(lr){2-5}
    & \multicolumn{2}{c}{SIMPLE} & \multicolumn{2}{c}{HumanoidArena} & Teacher \\
    \cmidrule(lr){2-3} \cmidrule(lr){4-5}
    Tracker & Hand & Walk & Hand & Walk & success \\
    \midrule
    HEFT & 0.9 & 11 & \textbf{0.7} & 8 & 94.2 \\
    \textbf{HEFT with harness data} & \textbf{0.6} & \textbf{5} & \textbf{0.7} & \textbf{3} & \textbf{97.9} \\
    SONIC & 2.6 & 35 & 2.0 & 44 & 27.9 \\
    AMO & 3.1 & 47 & 3.0 & 40 & 12.9 \\
    \midrule
    \multicolumn{6}{l}{\textit{HEFT with harness data, restricted commands}} \\
    No torso pitch or yaw & -- & -- & -- & -- & 67.1 \\
    No palm orientation & -- & -- & -- & -- & 18.8 \\
    \bottomrule
  \end{tabular}
\end{table}

We evaluate the execution interface with the teacher policies in place of the agent, so that the results reflect
execution and not the agent's decisions. We measure two things (\cref{tab:execution}). The first is
tracking error. Each tracker follows the same references, which the planner produced from the commands of a
fine-tuned agent in its SIMPLE and HumanoidArena evaluation episodes and on which no tracker was trained. We measure
the distance of each hand from its reference position relative to the pelvis, over the frames in which the arm moves,
and the error of each walk relative to the commanded distance. The second is task success: the teachers solve 240 episodes, half of the
base tasks and half of the SIMPLE tasks in our simulation, first with the tracker replaced and then with some
commands removed. We compare our tracker, HEFT further trained on harness data, the references that the planner
produced in teacher episodes of the base tasks, with HEFT before this training, with SONIC~\cite{luo2025sonic}, which follows the same whole-body references, and with
AMO~\cite{li2025amo}. AMO takes pelvis height, torso orientation, and walking velocity for its lower body and joint targets
for its arms, which we derive from the same references.

Our tracker executes the agent's commands most accurately. Its hands stay within \SI{0.7}{cm} of the reference, and
its walks end within 5\% of the commanded distance. Training on harness data mainly reduces the walk error,
from 11\% to 5\% on SIMPLE and from 8\% to 3\% on HumanoidArena. SONIC and AMO track the hands 2.0 to \SI{3.1}{cm}
from the reference, and their walk error depends mostly on the direction. SONIC follows forward walks closely,
covering 90\% of the commanded distance, and both turn in place within 4 to 13\% of the commanded angle. Side and
back steps are harder: SONIC covers 6\% of a side step and 39\% of a back step, and AMO about half of either. SONIC
also keeps its torso more upright than commanded, leaning \SI{11}{\degree} when the reference leans
\SI{34}{\degree}.

These differences decide task success. With our tracker, the teachers succeed in 97.9\% of the episodes, against
94.2\% with HEFT before training on harness data, 27.9\% with SONIC, and 12.9\% with AMO. The teachers line the robot
up with a table or a handle through short side and back steps, and lean over tables to reach. When such a step falls short, the robot stops far from where the teacher planned. The teacher then adjusts
its position again, but these corrections are again side and back steps, so they fail in the same way. Grasps,
pushes, and pulls then fail because their targets are out of reach.

The tasks also rely on the torso and palm commands. Without torso pitch and yaw, teacher success falls to 67.1\%:
picking from a low table and carrying a box fail in most episodes. Without palm orientation, it falls to 18.8\%,
since grasps and door handles need the palm turned toward them. The fine-tuned agent uses these commands as the teachers do: it leans or
turns its torso in 66\% of its SIMPLE episodes and 20\% of its HumanoidArena episodes, and sets the palm orientation
in 100\% and 75\%.

\subsection{Transfer from the Base Tasks}
\label{sec:exp-transfer}
\begin{table}[t]
  \caption{Success rate (\%) without reasoning, by the number $N$ of SIMPLE trajectories per task.}
  \label{tab:agent}
  \centering
  \begin{tabular}{lcccccc}
    \toprule
    & \multicolumn{4}{c}{SIMPLE} & \multicolumn{2}{c}{HumanoidArena} \\
    & & & & & \multicolumn{2}{c}{(subset)} \\
    \cmidrule(lr){2-5} \cmidrule(lr){6-7}
    $N$ & 0 & 5 & 10 & 30 & 0 & 10 \\
    \midrule
    Without base tasks & 3.3$^\dagger$ & 90.0 & 95.0 & 96.7 & 16.3$^\dagger$ & 18.8 \\
    With base tasks & \textbf{51.1} & \textbf{95.0} & \textbf{98.9} & \textbf{98.9} & \textbf{42.5} & \textbf{28.8} \\
    \midrule
    \multicolumn{7}{l}{\textit{Without perception}} \\
    With base tasks & 2.8 & -- & 80.6 & -- & 27.5 & 21.3 \\
    \bottomrule
  \end{tabular}\\[2pt]
  {\footnotesize $^\dagger$Pretrained model.}
\end{table}

Training on the base tasks alone raises success from 3.3\% to 51.1\% on SIMPLE and from 16.3\% to 42.5\% on
HumanoidArena (\cref{tab:agent}), although neither benchmark's tasks appear in its training data. The much larger
Qwen3.8-Max succeeds in only 12.2\% of SIMPLE's episodes without thinking and 7.8\% with it
(\cref{tab:reasoning}).

The transfer to HumanoidArena comes mainly from the base tasks. The model trained only on
ten SIMPLE trajectories per task succeeds in 95.0\% of SIMPLE's episodes but in only 18.8\% of HumanoidArena's, close to the pretrained model.
Adding the same SIMPLE trajectories to the base tasks raises SIMPLE success from 51.1\% to 98.9\% but lowers
HumanoidArena success from 42.5\% to 28.8\%. Every SIMPLE trajectory picks up an object from a table in front of the robot, and the agent leans
toward this pattern: on HumanoidArena, it succeeds more often at carrying a box from a table to a shelf, but less
often at the tasks that differ from it, such as navigation and opening a door.

\subsection{Learning New Tasks from a Few Trajectories}
\label{sec:exp-fewshot}

A few trajectories per task suffice for the agent to learn a SIMPLE task (\cref{tab:agent}). Five trajectories
per task alone raise success from 3.3\% to 90.0\%. Added to the base tasks, five trajectories per task reach 95.0\%,
and ten reach 98.9\%.

Training on the base tasks as well raises success at every number of trajectories. Among the models trained on
SIMPLE trajectories, the gain is largest at $N{=}5$, 95.0\% against 90.0\%, and narrows as both approach saturation.

\subsection{Reliance on Perception}
\label{sec:exp-perception}
To measure how much the agent relies on perception, we train models without it: we remove every perception call,
together with its returned points and image, from the training data and from the agent's interface. The agent still
sees each head-camera image, sets its gaze, and receives its body state, but it can no longer measure 3D positions.

Without perception, success drops on both benchmarks, and the drop grows with how much a task relies on precise
positions (\cref{tab:agent}, bottom). For the model trained on the base tasks only, SIMPLE success falls from
51.1\% to 2.8\%. Every SIMPLE task requires a grasp, and a palm placed off target often knocks the object over.
HumanoidArena success falls less, from 42.5\% to 27.5\%, since its tasks are mostly walking and only a few steps
need precise positions. With ten SIMPLE trajectories per task added, SIMPLE success still reaches 80.6\%, against
98.9\% with perception. Most SIMPLE tasks pick up the same cracker box, so the agent can learn from these
trajectories to judge where the box is from how it looks in the image.

\subsection{Effect of Reasoning}
\label{sec:exp-reasoning}
\begin{table}[t]
  \caption{Success rate (\%) without (w/o) and with (w/) reasoning.}
  \label{tab:reasoning}
  \centering
  \begin{tabular}{lcccc}
    \toprule
    & \multicolumn{2}{c}{SIMPLE} & \multicolumn{2}{c}{HumanoidArena} \\
    & & & \multicolumn{2}{c}{(subset)} \\
    \cmidrule(lr){2-3} \cmidrule(lr){4-5}
    Training data & w/o & w/ & w/o & w/ \\
    \midrule
    \multicolumn{5}{l}{\textit{Qwen3.8-27B}} \\
    None (pretrained) & 3.3 & 8.3 & 16.3 & 22.5 \\
    Base tasks & 51.1 & 47.2 & 42.5 & 41.3 \\
    SIMPLE ($N{=}10$) & 95.0 & 86.7 & 18.8 & 12.5 \\
    Base tasks + SIMPLE ($N{=}10$) & 98.9 & 89.4 & 28.8 & 38.8 \\
    \midrule
    \multicolumn{5}{l}{\textit{Qwen3.8-Max}} \\
    None (API) & 12.2 & 7.8 & -- & -- \\
    \bottomrule
  \end{tabular}
\end{table}

The effect of reasoning depends on the model and the benchmark (\cref{tab:reasoning}). The pretrained Qwen3.8-27B succeeds
more often with reasoning on both benchmarks. After fine-tuning, the models without reasoning succeed more often on
SIMPLE, while on HumanoidArena neither variant is consistently ahead.

On SIMPLE, three kinds of errors account for most failures with reasoning. The main one starts from a wrong reading:
the agent selects a point off the object, such as on the table behind it, and its reasoning computes the next targets
from this reading without questioning it, for example concluding that a box is several times deeper than it is. The
teachers almost never make errors in perception or reasoning, so the demonstrations never show how to notice and correct
such an error. The other two are an arithmetic error in the reasoning, made even when the readings are correct,
and reasoning that drifts into the steps of another task, which the actions then follow. Without reasoning, the agent keeps its
targets close to those of the demonstrations, which suits SIMPLE's similar scenes. Reasoning does help when a grasp
needs an offset from the measured point, which it often computes correctly. Demonstrations that include errors and
their corrections, such as a wrong reading that the reasoning detects and measures again, could teach the agent to
check its own reasoning and let reasoning help in scenes that vary more.

\subsection{Transfer to a Real G1}
\label{sec:exp-real}
We run the model trained on the base tasks, without reasoning, on a real G1. Its training data contain no real
images, and the harness is the same as in simulation except for the camera calibration. The
robot has rubber hands without finger control. With these hands
the robot can hold an object only between its palms, so we test two tasks: carrying a box with both
hands from one table to another, and walking around obstacles to a goal. The model runs on a remote server
and exchanges observations and commands with the robot over Wi-Fi. An operator judges each episode by the
conditions in its instruction: the box stands upright on the other table after both hands let go, or the robot
stops about \SI{0.5}{m} from the goal without touching it or an obstacle.

Besides real images and depth, the scenes differ from those in training: the white folding table is smaller and
the office desk is wider than any table in training, and the navigation goal is a stool, while every goal in
training is a colored mat on the floor. For carrying, the robot moves a box about \SI{20}{cm} on each side in two
mirrored layouts with ten episodes each: from the folding table to the desk on its right, and from the desk to the
folding table on its left. Between episodes, we vary the robot's starting position, the box's position, and the
orientation of the tables. For navigation, cardboard boxes stand between the robot and the stool, and between
episodes we vary the robot's starting position and the placement of the boxes.

The agent succeeds in 13 of 20 carrying episodes (8 of 10 toward the desk and 5 of 10 toward the folding table) and
in 6 of 10 navigation episodes (\cref{fig:filmstrips}).

\begin{table}[t]
  \caption{Time in successful real episodes (s) and its share (\%), mean per episode, from the first image until
    the motion that completes the task ends.}
  \label{tab:timing}
  \centering
  \begin{tabular}{lcccc}
    \toprule
    & \multicolumn{2}{c}{Carrying} & \multicolumn{2}{c}{Navigation} \\
    \cmidrule(lr){2-3} \cmidrule(lr){4-5}
    & s & \% & s & \% \\
    \midrule
    Perception & 12.1 & 20 & 5.4 & 20 \\
    \quad Image capture & 11.3 & 19 & 4.9 & 18 \\
    \quad Perception queries & 0.8 & 1 & 0.4 & 2 \\
    Decision (model requests) & 11.5 & 19 & 5.0 & 19 \\
    Execution & 37.0 & 61 & 16.4 & 61 \\
    \quad Planning & 6.2 & 10 & 3.0 & 11 \\
    \quad Sending and queueing & 4.0 & 7 & 1.4 & 5 \\
    \quad Motion & 26.8 & 44 & 12.0 & 45 \\
    \midrule
    Episode & 60.7 & 100 & 26.8 & 100 \\
    \bottomrule
  \end{tabular}
\end{table}

\Cref{tab:timing} splits each successful episode into perception, decision, and execution. Most of the time goes
to moving the robot. Decision takes about a fifth: each request is short without reasoning, but an episode makes
many of them. Image capture takes as long, mostly waiting for the head and body to settle, so faster capture and
planning would shorten episodes more than a faster model would.

\section{Conclusion}
\label{sec:conclusion}
We built HITCH, a harness through which a VLM controls a humanoid. Its execution interface turns body-part commands into
references through a planner, and a tracker trained further on these references executes them accurately enough
that teacher policies succeed in 97.9\% of episodes, against 27.9\% and 12.9\% with baseline trackers. Its
perception interface returns the 3D position of any point the model selects; without it, the model trained on the
base tasks solves only 2.8\% of SIMPLE episodes. Fine-tuned through this harness on teacher trajectories of five
base tasks, Qwen3.8-27B learns skills that carry over: it succeeds in 51.1\% of SIMPLE and 42.5\% of HumanoidArena
episodes without task-specific trajectories, reaches 95.0\% on SIMPLE with five per task, and operates a real G1
without real training data.

Several limitations remain. The planner does not support motions such as sitting or stepping over obstacles; a
motion generation model could produce their references for the same tracker. Adding SIMPLE trajectories lowers
HumanoidArena success, so data from one benchmark can narrow what the base tasks teach. After fine-tuning,
reasoning lowers SIMPLE success, mostly because the agent builds on wrong readings; demonstrations with such errors
and their corrections could teach it to check its own reasoning. Each command runs to completion before the agent
observes again, so the agent cannot react during a motion. On the real G1, hands without finger control limited us
to two tasks.

\bibliographystyle{IEEEtran}
\bibliography{refs}

\begin{thebibliography}{10}
\providecommand{\url}[1]{#1}
\csname url@samestyle\endcsname
\providecommand{\newblock}{\relax}
\providecommand{\bibinfo}[2]{#2}
\providecommand{\BIBentrySTDinterwordspacing}{\spaceskip=0pt\relax}
\providecommand{\BIBentryALTinterwordstretchfactor}{4}
\providecommand{\BIBentryALTinterwordspacing}{\spaceskip=\fontdimen2\font plus
\BIBentryALTinterwordstretchfactor\fontdimen3\font minus
  \fontdimen4\font\relax}
\providecommand{\BIBforeignlanguage}[2]{{%
\expandafter\ifx\csname l@#1\endcsname\relax
\typeout{** WARNING: IEEEtran.bst: No hyphenation pattern has been}%
\typeout{** loaded for the language `#1'. Using the pattern for}%
\typeout{** the default language instead.}%
\else
\language=\csname l@#1\endcsname
\fi
#2}}
\providecommand{\BIBdecl}{\relax}
\BIBdecl

\bibitem{wei2026psi0}
S.~Wei \emph{et~al.}, ``{$\Psi_0$: An Open Foundation Model Towards Universal
  Humanoid Loco-Manipulation},'' in \emph{Robotics: Science and Systems (RSS)},
  2026.

\bibitem{shao2026holom}
W.~Shao \emph{et~al.}, ``{Humanoid Loco-Manipulation With Discrete VLA
  Model},'' arXiv preprint arXiv:2609.35709, 2026.

\bibitem{yue2026beingm07}
J.~Yue \emph{et~al.}, ``{Being-M0.7: A Latent World-Action Model for Humanoid
  Robots},'' arXiv preprint arXiv:2610.11283, 2026.

\bibitem{wei2026simple}
S.~Wei \emph{et~al.}, ``{SIMPLE: Simulation-Based Policy Learning and
  Evaluation for Humanoid Loco-manipulation},'' in \emph{Conference on Robot
  Learning (CoRL)}, 2026.

\bibitem{liang2022codepolicies}
J.~Liang \emph{et~al.}, ``{Code as Policies: Language Model Programs for
  Embodied Control},'' in \emph{IEEE International Conference on Robotics and
  Automation (ICRA)}, 2023, pp. 9493--9500.

\bibitem{huang2023voxposer}
W.~Huang, C.~Wang, R.~Zhang, Y.~Li, J.~Wu, and L.~Fei-Fei, ``{VoxPoser:
  Composable 3D Value Maps for Robotic Manipulation with Language Models},'' in
  \emph{Conference on Robot Learning (CoRL)}, ser. Proceedings of Machine
  Learning Research, vol. 229, 2023, pp. 540--562.

\bibitem{huang2024rekep}
W.~Huang, C.~Wang, Y.~Li, R.~Zhang, and L.~Fei-Fei, ``{ReKep: Spatio-Temporal
  Reasoning of Relational Keypoint Constraints for Robotic Manipulation},'' in
  \emph{Conference on Robot Learning (CoRL)}, ser. Proceedings of Machine
  Learning Research, vol. 270, 2024, pp. 4573--4602.

\bibitem{yuan2025being0}
H.~Yuan \emph{et~al.}, ``{Being-0: A Humanoid Robotic Agent with
  Vision-Language Models and Modular Skills},'' arXiv preprint
  arXiv:2503.12533, 2025.

\bibitem{ren2026cybowaiter}
P.~Ren \emph{et~al.}, ``{Cybo-Waiter: A Physical Agentic Framework for Humanoid
  Whole-Body Locomotion-Manipulation},'' arXiv preprint arXiv:2603.10675, 2026.

\bibitem{huang2026rax}
R.~Huang \emph{et~al.}, ``{Scaling Online Exploration for Robot Agents},''
  arXiv preprint arXiv:2610.22496, 2026.

\bibitem{wang2026humanoidarena}
T.~Wang \emph{et~al.}, ``{HumanoidArena: Benchmarking Egocentric Hierarchical
  Whole-body Learning},'' in \emph{Advances in Neural Information Processing
  Systems (NeurIPS)}, 2026.

\bibitem{he2024omnih2o}
T.~He \emph{et~al.}, ``{OmniH2O: Universal and Dexterous Human-to-Humanoid
  Whole-Body Teleoperation and Learning},'' in \emph{Conference on Robot
  Learning (CoRL)}, ser. Proceedings of Machine Learning Research, vol. 270,
  2024, pp. 1516--1540.

\bibitem{ji2024exbody2}
M.~Ji \emph{et~al.}, ``{ExBody2: Advanced Expressive Humanoid Whole-Body
  Control},'' in \emph{IEEE International Conference on Robotics and Automation
  (ICRA)}, 2026, pp. 613--620.

\bibitem{chen2025gmt}
Z.~Chen, M.~Ji, X.~Cheng, X.~Peng, X.~B. Peng, and X.~Wang, ``{GMT: General
  Motion Tracking for Humanoid Whole-Body Control},'' arXiv preprint
  arXiv:2506.14770, 2025.

\bibitem{liao2025beyondmimic}
Q.~Liao \emph{et~al.}, ``{BeyondMimic: From Motion Tracking to Versatile
  Humanoid Control via Guided Diffusion},'' \emph{Science Robotics}, vol.~11,
  no. 117, p. eadx8924, 2026.

\bibitem{ze2025twist2}
Y.~Ze \emph{et~al.}, ``{TWIST2: Scalable, Portable, and Holistic Humanoid Data
  Collection System},'' in \emph{IEEE International Conference on Robotics and
  Automation (ICRA)}, 2026, pp. 15\,900--15\,907.

\bibitem{chen2026holomotion}
M.~Chen \emph{et~al.}, ``{HoloMotion-1 Technical Report},'' arXiv preprint
  arXiv:2605.15336, 2026.

\bibitem{liu2026heft}
C.~Liu \emph{et~al.}, ``{HEFT: Heavy-Payload Full-size Humanoid Teleoperation
  with Privileged Motion Guidance and Windowed Payload Curriculum},'' in
  \emph{Conference on Robot Learning (CoRL)}, 2026.

\bibitem{luo2025sonic}
Z.~Luo \emph{et~al.}, ``{SONIC: Supersizing Motion Tracking for Natural
  Humanoid Whole-Body Control},'' \emph{Science Robotics}, vol.~11, no. 117, p.
  eaed4592, 2026.

\bibitem{li2025amo}
J.~Li, X.~Cheng, T.~Huang, S.~Yang, R.-Z. Qiu, and X.~Wang, ``{AMO: Adaptive
  Motion Optimization for Hyper-Dexterous Humanoid Whole-Body Control},'' in
  \emph{Robotics: Science and Systems (RSS)}, 2025.

\bibitem{ben2025homie}
Q.~Ben, F.~Jia, J.~Zeng, J.~Dong, D.~Lin, and J.~Pang, ``{HOMIE: Humanoid
  Loco-Manipulation with Isomorphic Exoskeleton Cockpit},'' in \emph{Robotics:
  Science and Systems (RSS)}, 2025.

\bibitem{he2024hover}
T.~He \emph{et~al.}, ``{HOVER: Versatile Neural Whole-Body Controller for
  Humanoid Robots},'' in \emph{IEEE International Conference on Robotics and
  Automation (ICRA)}, 2025, pp. 9989--9996.

\bibitem{dong2026hero}
R.~Dong, Z.~Li, A.~Gupta, X.~He, and S.~Gupta, ``{HERO: Learning Humanoid
  End-Effector Control for Visual Whole-Body Open-Vocabulary Object
  Grasping},'' in \emph{Conference on Robot Learning (CoRL)}, 2026.

\bibitem{kim2024openvla}
M.~J. Kim \emph{et~al.}, ``{OpenVLA: An Open-Source Vision-Language-Action
  Model},'' in \emph{Conference on Robot Learning (CoRL)}, ser. Proceedings of
  Machine Learning Research, vol. 270, 2024, pp. 2679--2713.

\bibitem{black2024pi0}
K.~Black \emph{et~al.}, ``{$\pi_0$: A Vision-Language-Action Flow Model for
  General Robot Control},'' in \emph{Robotics: Science and Systems (RSS)},
  2025.

\bibitem{physicalintelligence2025pi05}
------, ``{$\pi_{0.5}$: a Vision-Language-Action Model with Open-World
  Generalization},'' in \emph{Conference on Robot Learning (CoRL)}, ser.
  Proceedings of Machine Learning Research, vol. 305, 2025, pp. 17--40.

\bibitem{ye2026dreamzero}
S.~Ye \emph{et~al.}, ``{World Action Models are Zero-shot Policies},'' in
  \emph{Conference on Robot Learning (CoRL)}, 2026.

\bibitem{nvidia2026cosmos3}
{NVIDIA} \emph{et~al.}, ``{Cosmos 3: Omnimodal World Models for Physical AI},''
  arXiv preprint arXiv:2606.02800, 2026.

\bibitem{xu2026lambda0}
{Lambda-humanoid Team} \emph{et~al.}, ``{$\lambda_0$: Scaling Open-World
  Egocentric Human Data for Humanoid Loco-Manipulation},'' arXiv preprint
  arXiv:2610.18950, 2026.

\bibitem{xue2025leverb}
H.~Xue \emph{et~al.}, ``{LeVERB: Humanoid Whole-Body Control with Latent
  Vision-Language Instruction},'' arXiv preprint arXiv:2506.13751, 2025.

\bibitem{cheng2024navila}
A.-C. Cheng \emph{et~al.}, ``{NaVILA: Legged Robot Vision-Language-Action Model
  for Navigation},'' in \emph{Robotics: Science and Systems (RSS)}, 2025.

\bibitem{bai2026egoactor}
Y.~Bai, M.~Yu, C.~Li, Z.~Bai, X.~Wang, and B.~F. Karlsson, ``{EgoActor:
  Grounding Task Planning into Spatial-aware Egocentric Actions for Humanoid
  Robots via Visual-Language Models},'' in \emph{Findings of the Association
  for Computational Linguistics: EMNLP 2026}, 2026, pp. 24\,059--24\,083.

\bibitem{ahn2022saycan}
B.~Ichter \emph{et~al.}, ``{Do As I Can, Not As I Say: Grounding Language in
  Robotic Affordances},'' in \emph{Conference on Robot Learning (CoRL)}, ser.
  Proceedings of Machine Learning Research, vol. 205, 2022, pp. 287--318.

\bibitem{huang2022innermonologue}
W.~Huang \emph{et~al.}, ``{Inner Monologue: Embodied Reasoning through Planning
  with Language Models},'' in \emph{Conference on Robot Learning (CoRL)}, ser.
  Proceedings of Machine Learning Research, vol. 205, 2022, pp. 1769--1782.

\bibitem{yang2026handoff}
L.~Yang \emph{et~al.}, ``{HANDOFF: Humanoid Agentic Task-Space Whole-Body
  Control via Distilled Complementary Teachers},'' arXiv preprint
  arXiv:2606.06493, 2026.

\bibitem{liu2026guava}
H.~Liu \emph{et~al.}, ``{Guava: Distilling Frontier VLMs into a Compact Agent
  through a Robotic Manipulation Harness},'' arXiv preprint arXiv:2606.18363,
  2026.

\bibitem{chen2026showharness}
Y.~Chen \emph{et~al.}, ``{Show-Harness: Just a VLM Agent Can Play Robots},''
  arXiv preprint arXiv:2609.10522, 2026.

\bibitem{weihs2020advisor}
L.~Weihs \emph{et~al.}, ``{Bridging the Imitation Gap by Adaptive
  Insubordination},'' in \emph{Advances in Neural Information Processing
  Systems (NeurIPS)}, vol.~34, 2021, pp. 19\,134--19\,146.

\bibitem{chen2024spatialvlm}
B.~Chen \emph{et~al.}, ``{SpatialVLM: Endowing Vision-Language Models with
  Spatial Reasoning Capabilities},'' in \emph{IEEE/CVF Conference on Computer
  Vision and Pattern Recognition (CVPR)}, 2024, pp. 14\,455--14\,465.

\bibitem{cheng2024spatialrgpt}
A.-C. Cheng \emph{et~al.}, ``{SpatialRGPT: Grounded Spatial Reasoning in Vision
  Language Models},'' in \emph{Advances in Neural Information Processing
  Systems (NeurIPS)}, vol.~37, 2024, pp. 135\,062--135\,093.

\bibitem{deitke2024molmo}
M.~Deitke \emph{et~al.}, ``{Molmo and PixMo: Open Weights and Open Data for
  State-of-the-Art Vision-Language Models},'' in \emph{IEEE/CVF Conference on
  Computer Vision and Pattern Recognition (CVPR)}, 2025, pp. 91--104.

\bibitem{zhou2025roborefer}
E.~Zhou \emph{et~al.}, ``{RoboRefer: Towards Spatial Referring with Reasoning
  in Vision-Language Models for Robotics},'' in \emph{Advances in Neural
  Information Processing Systems (NeurIPS)}, vol.~38, 2025, pp.
  28\,404--28\,481.

\bibitem{tan2026lingbotdepth}
B.~Tan \emph{et~al.}, ``{Masked Depth Modeling for Spatial Perception},'' in
  \emph{European Conference on Computer Vision (ECCV)}, ser. Lecture Notes in
  Computer Science, 2026, pp. 453--470.

\bibitem{bai2025qwen3vl}
S.~Bai \emph{et~al.}, ``{Qwen3-VL Technical Report},'' arXiv preprint
  arXiv:2511.21631, 2025.

\end{thebibliography}

\clearpage
\appendices

\section{Interface}
\label{app:interface}
\textbf{Commands.} \Cref{tab:fields} lists the fields of an execution command. All positions are in the
observation frame $O$ of the latest observation, in metres. A command executes its fields in the order of the table:
the body and hands move in place, the fingers close or open, the robot walks, and the camera turns last.

\begin{table}[!h]
  \caption{Fields of an execution command.}
  \label{tab:fields}
  \centering
  \footnotesize
  \begin{tabular}{lp{0.33\textwidth}}
    \toprule
    Field & Content \\
    \midrule
    Body & Pelvis height, \SI{0.30}{m} to \SI{0.79}{m}; torso pitch, up to \SI{60}{\degree}, and yaw, up to
      \SI{45}{\degree} either way, in steps of \SI{5}{\degree}, with limits that depend on the pelvis height;
      duration. \\
    Hands & Per hand, 1 to 12 palm waypoints, each with a position, an optional orientation, and a duration of
      \SI{0.4}{s} to \SI{8}{s}; or the hand stays fixed in space, moves with the pelvis or the torso, or returns to
      its default pose. An orientation is a named palm direction (inward, down, up, or forward) or a palm normal and
      a finger direction. \\
    Fingers & Close or open, on dexterous hands. \\
    Walking & A target position and turn in $O$, or 1 to 12 footsteps. \\
    Gaze & Camera yaw, \SI{-45}{\degree} to \SI{45}{\degree}, and pitch, \SI{0}{\degree} to \SI{60}{\degree}
      down, in steps of \SI{15}{\degree}. \\
    \bottomrule
  \end{tabular}
\end{table}

\begin{table*}[t]
  \caption{Official instructions and ours.}
  \label{tab:instructions}
  \centering
  \footnotesize
  \begin{tabular}{p{0.13\textwidth}p{0.31\textwidth}p{0.48\textwidth}}
    \toprule
    Task & Official & Ours \\
    \midrule
    \multicolumn{3}{l}{\textit{SIMPLE}} \\
    XMovePick & move forward to pick up the cracker box & Walk to the table ahead and pick up the cracker box with one
      hand. \\
    XMoveBendPick & xmove to the table and bend to pick up the cracker box & Walk to the low table ahead and pick up
      the cracker box with one hand. \\
    BendPick & bend to pick up the cracker box & Pick up the cracker box from the low table in front of you with one
      hand. \\
    Grasp & grasp the object on the tabletop. & Pick up the chips can from the table in front of you with one hand. \\
    PickBetweenTables & pick up the cracker box from table1,locomotion to table2,and place on table2. & Pick up the
      cracker box from the table in front of you, carry it to the basket on the table behind you, and put it into the
      basket. Use one hand. \\
    Handover & Hand over cracker box from right hand to left hand and place it on the container. & Pick up the cracker
      box with your right hand, hand it over to your left hand, and put it into the basket on your left. \\
    \midrule
    \multicolumn{3}{l}{\textit{HumanoidArena}} \\
    Navigation & Avoid obstacles and move to the yellow marked area. & Walk to the yellow wet-floor sign ahead of you
      and stop inside the yellow square taped on the floor around it, steering clear of the pink obstacles. End with
      both feet fully inside the yellow square. \\
    Boxing & Strike the green markers on the punching bag. & One of the two punching bags hanging ahead of you has a
      small green dot on its surface, possibly on the side facing away from you. Find the green dot and punch it:
      touch it with one fist. \\
    Open door & Open the door. & Open the door in front of you. Press its handle down to unlatch it; it opens away
      from you, so push it until it stands wide open. \\
    Box to shelf & Move the box from the table onto the shelf. & With both hands, pick up the brown box from the
      rectangular table in front of you and put it on the shelving unit to your left. Use a shelf that has enough
      room for it. Release both hands and check that the object stays upright and in place. \\
    \bottomrule
  \end{tabular}
\end{table*}

\textbf{Perception queries.} A query gives 1 to 100 points in the latest image as integer coordinates in
$[0, 1000]$. The reply lists the position of each point in $O$, marks points outside the image or without depth as
invalid, and warns when the depth around a point varies by more than \SI{3}{cm}. The returned image numbers the
points and colors the valid ones green and the invalid ones red.

\textbf{Example.} \Cref{fig:turns} shows two consecutive turns of a teacher episode of the pick-and-place base task,
in which the robot must put a cube onto a mat behind it. The teacher measures the cube and the counter under it, then
raises its right hand above the cube while it steps to the counter, and turns the camera down.

\begin{figure*}[t]
  \centering
  \scriptsize
\begin{verbatim}
Agent:   perceive {"points": [[602, 612], [650, 620], [631, 569], [630, 513], [543, 718], [723, 722],
                              [634, 767]]}
Harness: Located 7 points on obs-2: 7 with depth, 0 without. Check the marked image: if a point is
         not on what you meant, perceive it again.
         1. image point (602, 612) -> (0.56, -0.15, 0.67)   5. image point (543, 718) -> (0.49, -0.07, 0.62)
         2. image point (650, 620) -> (0.55, -0.21, 0.67)   6. image point (723, 722) -> (0.48, -0.30, 0.62)
         3. image point (631, 569) -> (0.57, -0.18, 0.71)   7. image point (634, 767) -> (0.46, -0.18, 0.60)
         4. image point (630, 513) -> (0.63, -0.19, 0.71)
         3D positions are in frame O of obs-2; they stay valid until the robot moves.  [marked image]
Agent:   execute {"hands": {"right": {"waypoints": [
           {"position": [0.18, -0.28, 0.82], "duration_s": 0.9,
            "orientation": {"normal": [-0.34, 0, -0.94], "fingers": [0.94, 0, -0.34]}},
           {"position": [0.33, -0.28, 0.82], "duration_s": 0.8,
            "orientation": {"normal": [-0.34, 0, -0.94], "fingers": [0.94, 0, -0.34]}}]}},
         "walk": {"target": [0.22, 0.01], "turn_deg": 0}, "eye": {"yaw_deg": 0, "pitch_deg": 60}}
Harness: It was executed to the end. Then the camera turned to yaw 0 deg, pitch 60 deg down from the torso.
         Observation obs-3. [...] The robot walked since obs-2: O moved with it [...].
         Coordinates from obs-2 no longer apply; measure again.
         - Body: standing; pelvis 0.75 m high (planned 0.75), roll 0 deg, pitch 0 deg. [...]
         - Right hand: palm (0.33, -0.28, 0.82) [...]; planned (0.33, -0.28, 0.82) [...];
           0.3 cm and 2 deg apart. Fingers open.
         [...]  [head-camera image]
\end{verbatim}
  \caption{Two turns of a teacher episode, shortened ([...]); the teacher's reasoning is omitted.}
  \label{fig:turns}
\end{figure*}

\section{Base Tasks}
\label{app:tasks}
Each episode of a base task draws a new scene. In every task, we also randomize the head-camera image (exposure,
gamma, white balance, contrast, saturation, blur, vignetting, and sensor noise) and the look of the room (lighting,
wall color, floor, and furniture textures); objects keep their colors, since the instructions refer to them by color.

\textbf{Pick and place.} The robot picks up one of one to three objects, cubes, balls, boxes, or cylinders of 50 to
\SI{300}{g}, from a table, round table, or counter whose top lies between \SI{0.40}{m} and \SI{0.90}{m}, and puts it
into a bin or bucket or onto a mat, on the same top or on another piece of furniture. In a quarter of the episodes, it
only picks up the object and holds it. It uses one hand. In 40\% of the episodes, the robot starts at the table;
otherwise, the table stands 0.85 to \SI{1.35}{m} ahead and the robot faces up to \SI{60}{\degree} away from it.

\textbf{Carrying.} The robot carries a box with a footprint of 0.18 to \SI{0.28}{m} with both hands from one piece of
furniture to another: tables, round tables, high stools and chairs, long counters, and shelving units, with tops
between \SI{0.40}{m} and \SI{0.77}{m}. The two pieces stand side
by side, at a corner, or facing each other, and the instruction gives the direction of the destination as one of
eight coarse bearings. In a quarter of the episodes, the robot only picks up the box and holds it.

\textbf{Navigation.} The robot walks onto a mat 2.5 to \SI{6.5}{m} away, named by its color and shape and sometimes
by a flag, sign, cone, pillar, or ball beside it, among up to three other mats. Up to three structures, such as
partitions, slaloms, and clusters of blocks, stand across the way, with more obstacles around them, and the robot
starts facing up to \SI{75}{\degree} away from the goal.

\textbf{Drawers and cabinet doors.} The robot opens or closes one drawer or door of a cabinet with one to three
columns of drawers, doors, or both, standing on a plinth or on legs. The instruction names the front by its row and
column. Doors are hinged on the left, the right, or the bottom, the handle of the target lies between \SI{0.70}{m}
and \SI{1.05}{m}, and a front to be closed starts partly open.

\textbf{Room doors.} The robot opens a door by pressing its lever handle down and pushing it, or closes a door that
stands 20 to \SI{35}{\degree} open. Doors are 0.80 to \SI{0.90}{m} wide and hinged on either side, the lever lies
between \SI{0.95}{m} and \SI{1.05}{m}, and the robot starts 1.4 to \SI{2.0}{m} away.

\section{Training}
\label{app:training}
All models start from the pretrained Qwen3.8-27B. We update all weights of the language model and keep the vision
encoder and its projector frozen. Every model uses AdamW with $\beta = (0.9, 0.95)$, a weight decay of 0.01, and
gradient clipping at 1.0; a learning rate of $5\times10^{-6}$ with 3\% warm-up and cosine decay to zero; a global
batch of 16 trajectories; bf16; and a maximum length of 49{,}152 tokens. Depending on the task, a trajectory has a
median of 14 to 33 turns and about 20k to 40k tokens, including the images.

The number of steps depends on the training data. A model trained on the base tasks only runs for 450 steps. A model
trained on SIMPLE trajectories only runs for 100 steps, and we keep its checkpoints at 60, 80, and 100 steps. A
model trained on the base tasks and SIMPLE trajectories runs for 750 steps, draws each trajectory from SIMPLE with
probability 0.1 (0.15 for $N{=}30$), and we keep its checkpoints at 450, 600, and 750 steps. For each model with
several checkpoints, we report the checkpoint with the highest success on SIMPLE.

\section{Execution Experiments}
\label{app:execution}
\textbf{Tracking error.} The references are those the planner streamed to the tracker in the evaluation episodes of
a fine-tuned agent: 180 SIMPLE episodes and 80 HumanoidArena episodes. Each tracker follows them in the same
simulator, on a G1 with rubber hands, without replanning. A step is the motion of one command. We measure the robot
and the reference from their own poses at the start of each step and compare the displacement of the pelvis over the
step, counting walks of at least \SI{0.2}{m}; the walk error is the distance between the executed and the planned
displacement, divided by the planned distance. The hand error is the distance between the wrist and its reference
position in the pelvis frame. It is averaged over the frames in which the reference hand is at least \SI{5}{cm} from
its position in the default stance, measured in the reference torso frame; in the other frames the arm rests at
its default pose.

\textbf{Trackers.} HEFT is the tracker before training on planner references, with the same architecture and
inputs as ours. SONIC is the released model, connected through the adapter of the sim2real tracking leaderboard
(\url{https://github.com/EGalahad/sim2real}), and follows the same whole-body references. AMO commands its lower body
through pelvis height, torso orientation, planar velocity, and heading, and its arms through joint targets. We derive
these commands from the reference at every control step: the height, torso orientation, and heading from the
reference pose, the velocity from the reference pelvis motion, and the arm and wrist joints from the reference
joints. We run AMO at its \SI{2}{ms} training step and let it step for any commanded velocity or heading, since
its release steps only at a forward speed of at least \SI{0.1}{m/s} and cannot turn in place.

\textbf{Teacher success.} The teachers solve 20 episodes of each of twelve tasks: the five base tasks, with picking and
placing split into a high-table and a low-table task, and the six SIMPLE tasks on the layouts of SIMPLE's
demonstrations. Without torso pitch and yaw, every
command keeps the torso at the default posture for its pelvis height. Without palm orientation, every hand waypoint
keeps only its position, and the planner chooses the orientation.

\section{Instructions}
\label{app:instructions}
\Cref{tab:instructions} lists the official instruction of each benchmark task and the instruction we use. The
official SIMPLE instructions are those of its released evaluation episodes, and the official HumanoidArena
instructions are those its evaluation server gives the policy.

\section{Time Limits and Episode Durations}
\label{app:durations}
\Cref{tab:durations} gives the official time limit of each benchmark task and how long our agent takes in its
successful episodes: on SIMPLE, the model trained on the base tasks and ten SIMPLE trajectories per task; on
HumanoidArena, the model trained on the base tasks only; both without reasoning. Times are simulated seconds from the
agent's first image to the first time the benchmark's check passes, and turns are the calls made up to that point.
The official limits match the speed of the policies these benchmarks evaluate, which act at high frequency. Our agent moves through a sequence of commands and
observes the scene between them, so it takes longer, and we allow \SI{45}{s} on SIMPLE and \SI{60}{s} on
HumanoidArena. A successful episode uses on average 4.8 perception calls and 10.3 commands on SIMPLE, and 6.8 and
10.6 on HumanoidArena. BendPick takes longer than XMoveBendPick, although the robot starts closer to its table: SIMPLE
starts it right at the low table, with the box at the table's near edge and the edge below the camera's view. The agent
first steps back to see the edge and set its stance from it, and then steps forward to grasp, while in XMoveBendPick it
sees the table from the start and walks to it directly.

\

\begin{table}[t]
  \caption{Time limits and durations of successful episodes. Time: mean and the 10th to 90th percentile; turns: mean
    and range.}
  \label{tab:durations}
  \centering
  \footnotesize
  \setlength{\tabcolsep}{4pt}
  \begin{tabular}{lcccc}
    \toprule
    & Official & \multicolumn{2}{c}{Time (s)} & Turns \\
    \cmidrule(lr){3-4}
    Task & limit (s) & Mean & 10th--90th & Mean (range) \\
    \midrule
    \multicolumn{5}{l}{\textit{SIMPLE}} \\
    XMovePick & 16 & 12.3 & 12.0--13.4 & 11.8 (9--13) \\
    XMoveBendPick & 16 & 14.0 & 13.9--14.0 & 11.5 (10--15) \\
    BendPick & 8 & 16.3 & 16.1--16.3 & 14.0 (11--17) \\
    Grasp & 27 & 14.2 & 14.1--14.4 & 11.0 (10--13) \\
    PickBetweenTables & 24 & 31.6 & 30.7--32.0 & 20.9 (20--22) \\
    Handover & 16 & 31.4 & 29.8--38.0 & 21.6 (15--26) \\
    All & & 20.0 & 12.0--31.7 & 15.2 (9--26) \\
    \midrule
    \multicolumn{5}{l}{\textit{HumanoidArena}} \\
    Navigation & 36 & 31.7 & 30.2--32.8 & 22.3 (14--26) \\
    Boxing & 18 & 17.3 & 10.9--20.9 & 12.5 (6--28) \\
    Open door & 36 & 32.0 & 30.2--34.5 & 20.3 (18--22) \\
    Box to shelf & 29 & 34.2 & 26.8--44.2 & 20.4 (14--27) \\
    All & & 26.6 & 13.5--36.8 & 17.4 (6--28) \\
    \bottomrule
  \end{tabular}
\end{table}

\end{document}